\documentclass{article}

\PassOptionsToPackage{numbers, compress}{natbib}

\usepackage[preprint]{neurips_2023}

\usepackage[utf8]{inputenc} 
\usepackage[T1]{fontenc}    
\usepackage{url}            
\usepackage{booktabs}       
\usepackage{amsfonts}       
\usepackage{nicefrac}       
\usepackage{microtype}      

\usepackage{amsmath,amsfonts,bm,amsthm}

\usepackage{thmtools,thm-restate}

\def\eqref#1{equation~(\ref{#1})}

\def\1{\bm{1}}

\DeclareMathAlphabet{\mathsfit}{\encodingdefault}{\sfdefault}{m}{sl}
\SetMathAlphabet{\mathsfit}{bold}{\encodingdefault}{\sfdefault}{bx}{n}

\usepackage{natbib}
\usepackage{tikz}
\usetikzlibrary{positioning}
\usepackage{subcaption}
\usepackage{hyperref}
\usepackage{url}
\usepackage{graphicx}
\usepackage{xcolor}
\usepackage{enumerate}
\usepackage{enumitem}
\usepackage{textcomp}
\usepackage{mathrsfs}
\usepackage{amsmath}
\usepackage{amssymb}
\usepackage{mathtools}
\usepackage{bbm}
\usepackage{epigraph}
\usepackage{etoolbox}
\usepackage{algorithm}
\usepackage{algpseudocode}
\usepackage{thmtools, thm-restate}
\usepackage{proof-at-the-end}
\usepackage{booktabs}
\usepackage{wrapfig}

\def\1{\mathbf{1}}

\def\1{\mathbf{1}}

\def\hat{\widehat}

\newcommand{\Sc}{\mathcal{S}}

\newcommand{\Dc}{\mathcal{D}}

\newcommand{\Lc}{\mathcal{L}}

\newcommand{\mc}[1]{\mathcal{#1}}

\usepackage{xcolor}
\usepackage{latexcolors}
\usepackage{hyperref}
\definecolor{stanfordred}{rgb}{0.54901961, 0.08235294, 0.08235294}
\usepackage[mathscr]{euscript}
\hypersetup{
    colorlinks=true,
    linkcolor=blue,
    filecolor=magenta,      
    urlcolor=purple,
    citecolor = stanfordred,
}

\newcount\Comments
\newcommand{\kibitz}[2]{\ifnum\Comments=1{\textcolor{#1}{\textsf{\footnotesize #2}}}\fi}
\usepackage{color}
\definecolor{darkred}{rgb}{0.7,0,0}
\definecolor{darkgreen}{rgb}{0.0,0.5,0.0}
\definecolor{darkblue}{rgb}{0.0,0.0,0.5}
\definecolor{teal}{rgb}{0.0,0.5,0.5}

\newcommand{\kl}[2]{D_{\mathrm{KL}}\left(#1 \mid\mid #2\right)}

\title{Shattering the Agent-Environment Interface for Fine-Tuning Inclusive Language Models}

\author{%
  Wanqiao Xu\\
  Department of Management Science \& Engineering\\
  Stanford University\\
  \texttt{wanqiaoxu@stanford.edu}\\
  \And
  Shi Dong\\
  Microsoft\\
  \texttt{dongshi@microsoft.com}\\
  \AND
  Dilip Arumugam\\
  Department of Computer Science\\
  Stanford University\\
  \texttt{dilip@cs.stanford.edu}\\
  \And
  Benjamin Van Roy \\
  Department of Electrical Engineering\\
  Department of Management Science \& Engineering\\
  Stanford University\\
  \texttt{bvr@stanford.edu}
}

\begin{document}

\maketitle


\begin{abstract}
    A centerpiece of the ever-popular reinforcement learning from human feedback (RLHF) approach to fine-tuning autoregressive language models is the explicit training of a reward model to emulate human feedback, distinct from the language model itself. This reward model is then coupled with policy-gradient methods to dramatically improve the alignment between language model outputs and desired responses. In this work, we adopt a novel perspective wherein a pre-trained language model is itself simultaneously a policy, reward function, and transition function. An immediate consequence of this is that reward learning and language model fine-tuning can be performed jointly and directly, without requiring any further downstream policy optimization. While this perspective does indeed break the traditional agent-environment interface, we nevertheless maintain that there can be enormous statistical benefits afforded by bringing to bear traditional algorithmic concepts from reinforcement learning. Our experiments demonstrate one concrete instance of this through efficient exploration based on the representation and resolution of epistemic uncertainty. In order to illustrate these ideas in a transparent manner, we restrict attention to a simple didactic data generating process and leave for future work extension to systems of practical scale.
\end{abstract}

\section{Introduction}

While recent years have witnessed a dramatic shift in the capabilities of generative AIs across numerous data modalities, excitement and discourse surrounding natural language processing (NLP) and large language models (LLMs) in particular has become near-ubiquitous within just the last few months~\citep{openai2023gpt4,touvron2023llama}, leading to an unprecedented proliferation of daily users probing and exploring these models' impressive capabilities through prolonged, interactive dialogues. With this attention has also come an onslaught of challenges for the AI and machine learning research communities, ranging from the rigorous benchmarking of capabilities~\citep{liang2022holistic}, adherence to copyright law~\citep{henderson2023foundation}, concerns for privacy~\citep{li2022large,li2022does}, and insight into the key methodologies for training these models~\citep{openai2023gpt4}, to name just a few. One question that lies at the heart of the last issue revolves around how much the successes of fine-tuned, autoregressive LLMs are driven by the reinforcement learning from human feedback (RLHF)~\citep{stiennon2020learning,ouyang2022training} pipeline?

While the classic NLP task of language modeling is easily formulated and solved through traditional supervised-learning techniques~\citep{bengio2000neural,mikolov2013efficient}, the RLHF paradigm has found great empirical success by interpreting this as merely a preliminary pretraining phase and further incorporating a subsequent fine-tuning phase that leverages human feedback when refining responses to be more accurate and more preferable. From the perspective of a sequential decision-making process, two hallmark characteristics of this pipeline include \textbf{(1)} viewing a language model as a policy, mapping a rich, pretrained representation of a sequence of tokens along with a partial response to a next-token distribution and \textbf{(2)} interpreting human feedback as identifying a terminal reward function that assigns scalar feedback to completed prompt-response pairs so as to incentivize preferred responses. In this work, we introduce a novel perspective on LLMs that extends \textbf{(1)} and renders \textbf{(2)} moot, giving rise to a novel and statistically-efficient fine-tuning method. 

We recognize that by virtue of vast amounts of unstructured Web data, a pretrained LLM can be simultaneously viewed as a policy, a reward function, and an environment simulator. Traditionally, a policy is implemented within a decision-making agent whereas the reward function and simulator are properties of the environment and, therefore, reside external to the agent.  Thus, our novel perspective blurs the traditional boundary between agent and environment found throughout the reinforcement-learning literature~\citep{sutton1998introduction}. Nevertheless, in this paper we demonstrate the value of this triumvirate through a meticulous collection of simple yet illustrative experiments, designed to highlight how foundational concepts from reinforcement learning can still be successfully brought to bear for fine-tuning LLMs. 

Concretely, through the lens of viewing a pretrained LLM as a reward function, we propose a new fine-tuning algorithm, Inclusive Learning From Human Feedback (ILHF), that offers two key advantages over current RLHF approaches. Firstly, from a computational perspective, ILHF avoids the need for further downstream application of policy-gradient methods~\citep{schulman2017proximal} in order to align LLM responses to human preferences. Secondly, from a statistical perspective and as our method's name suggests, the LLMs resulting from ILHF are \textit{inclusive}~\citep{arumugam2022inclusive} and, therefore, demonstrably converge to the preferred population response distribution over the course of fine-tuning; this stands in stark contrast to the \textit{agglomerative} models that arise from the standard RLHF approach, which are encouraged to place all probability mass on a singular, ``best'' response that is preferred by the majority of the population. Beyond empirical results that validate the emergence of such inclusive and agglomerative models, we further demonstrate how ILHF, a supervised-learning approach, can still be made more statistically efficient by leveraging judicious exploration strategies borne out in the reinforcement-learning literature~\citep{lu2017ensemble}.

The paper proceeds as follows: in Section \ref{sec:prelims} we establish notation, review the current RLHF pipeline, and briefly present empirical results on a didactic example to highlight the difference between inclusive and agglomerative LLMs. We then proceed to outline ILHF in Section \ref{sec:approach} followed by details of our experimental protocol in Section \ref{sec:exps} and a discussion of empirical results in Section \ref{sec:disc}.

\section{Preliminaries}
\label{sec:prelims}

\subsection{Notation}

For any natural number $N \in \mathbb{N}$, we denote the index set as $[N] \triangleq \{1,2,\ldots,N\}$. For any arbitrary set $\mc{Z}$, we use the Kleene plus $\mc{Z}^+$ to denote the set of all sequences of length at least one formed by elements of $\mc{Z}$. Orthogonally, we use  $\Delta(\mc{Z})$ to denote the set of all probability distributions supported on $\mc{Z}$. At the most abstract level, a LLM is an autoregressive mapping that, given a current sequence of tokens, generates a probability distribution over next tokens; modern applications of LLMs often elide this low-level mechanistic view of these models and instead adopt the more holistic perspective that a LLM maps an input prompt to a response, both of which are variable-length sequences of tokens. If $\mathcal{V}$ is the vocabulary or set of all possible tokens, then a LLM is represented as a mapping $\pi_\phi: \mc{V}^+ \rightarrow \Delta(\mc{V})$ parameterized by $\phi \in \Re^d$ where some initial, non-empty sequence of tokens from $\mc{V}$ (constituting a prompt) and subsequently sampled tokens for the response are autoregressively passed back into $\pi$ as inputs to generate the next-token distribution. With a slight abuse of notation, we use $\overline{\pi}_{\phi}: \mc{V}^+ \rightarrow \Delta(\mc{V}^+)$ to denote the associated mapping from an input prompt to a distribution over full, complete responses.
 
\subsection{Reinforcement Learning from Human Feedback (RLHF)}

Current approaches to RLHF~\citep{stiennon2020learning,ouyang2022training} are characterized by three distinct phases: pretraining, reward model learning, and fine-tuning. Pretraining is facilitated by curating a large dataset of $N \in \mathbb{N}$ prompt-response pairs $\mathcal{D} = \{(\overline{X}_i,\overline{Y}_i)\}_{i=1}^N$ where $\overline{X}_i, \overline{Y}_i \in \mc{V}^+$, typically representing unstructured text data scraped from the Web. Pretrained LLM parameters $\phi^{\mathrm{pre}} \in \Re^d$ are obtained through the standard supervised language-modeling objective which, in this context, aligns with the classic behavioral cloning~\citep{bain1999framework,pomerleau1989alvinn} loss function used widely in imitation learning: 
$\mathcal{L}^{\mathrm{pre}}(\phi) = - \frac{1}{N} \sum\limits_{i=1}^N  \sum\limits_{j=1}^{\lambda(i)} \log\left(\overline{\pi}_{\phi}(\overline{Y}_{i,j} \mid \overline{X}_i, \overline{Y}_{i,1:j-1})\right),$
where $\lambda(i)$ denotes the length of the $i$th response, $\overline{Y}_i$. The challenge posed after the completion of pretraining is that the unstructured text data curated in $\mc{D}$ is only an approximation to proper, natural text data that end users want and expect from a LLM; this is a direct consequence of quickly collating $\mc{D}$ by scraping the Internet. Moreover, beyond these initial syntactic issues, such Web sources are also fraught with errors and factual inaccuracies that need to be corrected as well. Oftentimes, these errors can be easily identified and remedied by human evaluators though the challenge lies in propagating such corrections completely throughout the vast space of possible prompts and responses.

Since the acquisition of feedback is the limiting reactant that inhibits scalability, a reward model is trained over the course of $H \in \mathbb{N}$ rounds to emulate human feedback obtained by iteratively fetching a single prompt $X_i$, sampling two random responses $Y_i^A,Y_i^B \sim \overline{\pi}_{\phi^{\mathrm{pre}}}(\cdot \mid X_i)$, and then querying a human evaluator for a binary indicator $L_i \in \{0,1\}$ that communicates their preference (or lack thereof) for the first response $Y_i^A$. An external reward model $r_\psi: \mc{V}^+ \times \mc{V}^+ \rightarrow \Re$ parameterized by $\psi \in \Re^m$ can then be trained via supervised learning by minimizing $\mc{L}^{\mathrm{reward}}(\psi) = -\frac{1}{H} \sum\limits_{i=1}^H \mathrm{LogSigmoid}\Big( R_\psi\big(X, Y_i^A, Y_i^B, L_i\big)\Big),$
where 
\[
    R_\psi\big(X, Y_i^A, Y_i^B, L_i\big) = 
    \begin{cases}
        r_\psi(X, Y_i^A) - r_\psi(X, Y_i^B) & \text{if $L_i = 1$}\\
        r_\psi(X, Y_i^B) - r_\psi(X, Y_i^A) & \text{otherwise}
    \end{cases}.
\]

With the fully-trained reward model $r_{\psi^\star}$ in hand, subsequent prompts can have their corresponding responses aligned to human preferences via reinforcement learning but without the need for laboriously querying a live human labeler. Specifically, for each of $T \in \mathbb{N}$ fine-tuning prompts $X_1,\ldots,X_T$, one can sample two responses $Y_{t,A}, Y_{t,B} \sim \overline{\pi}_{\phi}(\cdot \mid X_t)$, obtain a synthetic human feedback signal $L_t(A,B)$ based on $r_{\psi^\star}(X_t,Y_{t,A})$ as well as $r_{\psi^\star}(X_t,Y_{t,B})$, and apply policy-gradient methods~\citep{williams1992simple,sutton1999policy,konda1999actor,mnih2016asynchronous,schulman2017proximal} in order to maximize $\mathcal{J}(\phi) = \mathbb{E}_{X_t, Y_{t,A}, Y_{t,B}}\left[ L_t(A,B)\right], \forall t \in [T]$. Naturally, as this is an objective for fine-tuning the LLM, initial policy parameters are set to be $\phi^{\mathrm{pre}}$. The default standard choice for carrying out this policy optimization in the existing literature is Proximal Policy Optimization (PPO)~\citep{schulman2017proximal}.

\subsection{Inclusive vs. Agglomerative AIs}
\label{subsec:inclusive}

\begin{wrapfigure}{R}{0.5\textwidth}
\centering
\includegraphics[width=0.45\textwidth]{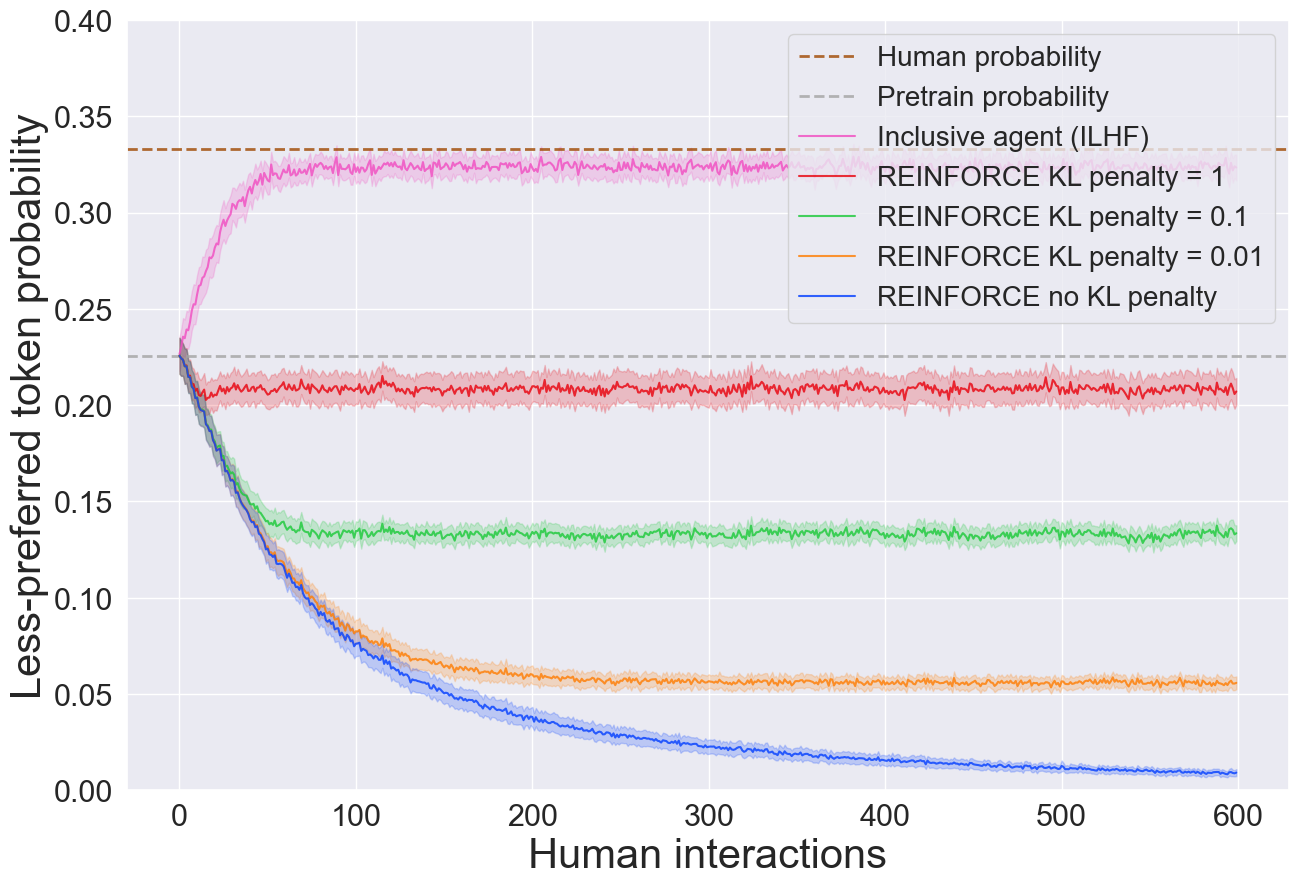}
\caption{ILHF succeeds and REINFORCE+KL penalty fails to match human response probability.}
\label{fig:toy-reinforce-rate}
\vspace{-0.5cm}
\end{wrapfigure}

By design, a LLM trained via the RLHF pipeline as outlined in the previous section learns to emit responses that maximize the likelihood of being preferred by human evaluators. Consequently, a RLHF model generating responses $Y_{t,A}$ across $T$ evaluation prompts $X_1,\ldots,X_T$ would likely be preferred (or be designated at least as good) as the alternative responses $Y_{t,B}$ of some other model $B$ according to the criterion $\sum\limits_{t=1}^T L_t(A,B)$. Unfortunately, as discussed and analyzed in \citet{arumugam2022inclusive}, LLMs that are preferred under this criterion are \textit{agglomerative} and allow for per-prompt response distributions that place all probability mass on a hypothetical ``best'' response which is simply preferred by the majority of human evaluators sampled from a given population (see Theorem 2 of \citet{arumugam2022inclusive}). Without delving into the details of the theoretical argument for this result, a simple intuition is that the fine-tuning phase of the RLHF pipeline operates as a contextual bandit~\citep{lattimore2020bandit}, for which there always exists an optimal policy that is a Dirac delta distribution on the optimal arm with highest expected mean reward, for each context. Not only does such a degenerate response distribution qualitatively fail to reflect the diverse interests and preferences of the overall population, but it also quantitatively precludes further downstream gradient-based optimization to redress the issue or cater to any shifts in the desired response distribution altogether. In contrast to an agglomerative model, one might instead favor an inclusive model that strives to preserve the full population response distribution. The primary contribution of this paper is a fine-tuning algorithm that leverages feedback signals derived from this preferred response distribution to yield such inclusive models.

To concretize the issues surrounding agglomerative models and to verify that such problems do manifest from current RLHF practice, we report results for a toy experiment using an extension of the simple, didactic example discussed in Section 2.2 of \citet{arumugam2022inclusive}. The example consists of multiple rounds of fine-tuning performed on exactly one prompt and where a response is a single token from $\{-1,1\}$. Initially, the model was pretrained to emit $-1$ with probability 0.77 and $1$ with probability 0.23. For fine-tuning, the desired population response distribution prefers response $-1$ with probability $\frac{2}{3}$ while response $1$ is favored with probability $\frac{1}{3}$.

For such a small-scale and simple problem, we capture the essence of the RLHF fine-tuning process through policy search by using REINFORCE~\citep{williams1992simple}, rather than PPO, along with KL-regularization towards the pretrained model response distribution; this implies that we explicitly forego the benefits of variance reduction that come from a critic as well as the potential for faster convergence through off-policy policy gradient updates. Results are provided in Figure \ref{fig:toy-reinforce-rate} varying the value of the KL-regularization coefficient. The primary observation is that as the REINFORCE fine-tuning updates induce an agglomerative model that emits the preferred token near-deterministically, the less-preferred token probability decays under the fine-tuned response distribution. Moreover, since the initial pretraining distribution underestimates the preference of the less-desired token, intensifying the KL-regularization can only halt this undesired behavior by pinning the model to the pretraining response distribution, but cannot otherwise alleviate the issue and recover the desired response distribution. In the next section, we introduce our ILHF fine-tuning approach that is also pictured in Figure \ref{fig:toy-reinforce-rate} and learns an inclusive model to emit the less-preferred token with the correct probability.


\section{Approach}
\label{sec:approach}

In this section, we outline the precise manner in which LLMs violate the agent-environment interface typically observed throughout the reinforcement-learning literature before introducing an alternative approach to RLHF fine-tuning that obviates the need for separate reward learning and policy optimization phases. While the resulting ILHF optimization does constitute a supervised learning problem, we proceed to introduce an augmented fine-tuning procedure that leverages solution concepts for efficient exploration in reinforcement learning.

\subsection{Shattering the Agent-Environment Interface}

We may formalize the sequential decision-making problem encapsulated by a LLM as a finite-horizon, episodic Markov Decision Process (MDP)~\citep{bellman1957markovian,Puterman94} defined by $\mc{M} = \langle \mc{S}, \mc{A}, \mc{R}, \mc{T}, \beta, H \rangle$. Specializing these MDP components to the language modeling problem, we may observe that the state space $\mc{S} = \mc{V}^+$ represents a sampled prompt as well as the current response generated thus far, the action space $\mc{A} = \mc{V}$ is the vocabulary of all possible tokens the LLM may generate, the initial state distribution $\beta \in \Delta(\mc{V}^+)$ represents an arbitrary distribution over prompts whose responses are adjusted over the course of the fine-tuning process, and the horizon $H \in \mathbb{N}$ is the maximum allowed response length which still enables variable-length responses shorter than $H$, akin to an episode of an MDP where the agent transitions to an absorbing terminal state before exhausting all $H$ steps. Naturally, a particular LLM with parameters $\phi \in \Re^d$ embodies a policy of this MDP $\pi_\phi: \mc{S} \rightarrow \Delta(\mc{A})$. All that remains is to define the reward function $\mc{R}:\mc{S} \times \mc{A} \rightarrow \Re$ providing evaluative feedback signals to the agent and the deterministic transition function $\mc{T}:\mc{S} \times \mc{A} \rightarrow \mc{S}$ yielding next states for each state-action pair.

Notably, the mechanics by which a single episode unfolds in this MDP violates the standard agent-environment interface, as the agent itself is a simulator that can sample rollouts for any given prompt. At the start of each episode, a new prompt is sampled $s_1 \sim \beta$ and, for each timestep $h \in [H]$, a LLM samples a next token $a_h \sim \pi_\phi(\cdot \mid s_h)$ before appending it to the current response yielding a deterministic next state $s_{h+1} = \mc{T}(s_h, a_h)$. More importantly, the fine-tuning approach introduced in the next section capitalizes on the realization that a suitable reward function for MDP $\mc{M}$ can be induced directly from the policy itself as $\mc{R}_{\phi}(s,a) = \log\left(\pi_\phi(a \mid s)\right)$. This again breaks the standard interface whereby rewards are computed within the confines of the environment and direct updates to policy parameters $\phi$ do not explicitly change the underlying reward function. While the idea of inducing a reward function (or, more generally, a cumulant~\citep{sutton2011horde}) from a policy and vice versa is not new~\citep{barreto2019option}, the implications for LLMs in particular stand to be quite profound; namely, it establishes a direct relationship between reward learning and policy optimization that the current RLHF paradigm segregates into distinct phases. In the next section, we provide a novel fine-tuning algorithm that leverages the equivalence between reward learning and policy optimization to consolidate these latter two stages of the RLHF pipeline.

\subsection{ILHF: A New Fine-Tuning Algorithm}

The previous section sets the stage for interpreting the output of the LLM pretraining phase as producing reward function parameters $\phi^{\mathrm{pre}}$ which analogously function as policy parameters. As pretraining typically occurs with a dataset that represents a crude approximation to proper written language (such as text scraped widely from the Internet), the corresponding reward function $\mc{R}_{\phi^{\mathrm{pre}}}(s,a) = \log\left(\pi_{\phi^{\mathrm{pre}}}(a \mid s)\right)$ is misspecified and the associated reward-maximizing policy $\overline{\pi}_{\phi^{\mathrm{pre}}}$ does not accurately reflect the desired response distribution. This begets the need for a loss function that refines reward function parameters to more accurately depict response preferences and, in doing so, refine the LLM policy parameters to induce a response distribution reflective of those preferences.

To that end, we offer the following loss function for optimizing reward function parameters and refining the LLM response distribution jointly. For any sampled prompt $X \sim \beta$, denote two i.i.d. sampled responses as $Y_i^A,Y_i^B \sim \overline{\pi}_{\phi}(\cdot \mid X)$ which are judged by a human evaluator according to $L_i \in \{0,1\}$. Define the binary probability distribution $\mc{P}_i = \begin{bmatrix} L_i, & 1-L_i\end{bmatrix}$ induced from the human evaluator. Then, we may induce a complementary distribution over the two sampled LLM responses as
\begin{align*}
    \mc{Q}_i^\phi &= \text{Softmax}\left(\left[\mc{R}_{\phi}(X,Y_i^A), \mc{R}_{\phi}(X,Y_i^B)\right]\right) = \begin{bmatrix} \frac{\pi_{\phi}(Y_i^A \mid X)}{\pi_{\phi}(Y_i^A \mid X) + \pi_{\phi}(Y_i^B \mid X)}, & \frac{\pi_{\phi}(Y_i^B \mid X)}{\pi_{\phi}(Y_i^A \mid X) + \pi_{\phi}(Y_i^B \mid X)}\end{bmatrix},
\end{align*}
in accordance with the Bradley-Terry model for pairwise comparisons~\citep{bradley1952rank}. Then, our proposed fine-tuning loss aims to minimize the KL-divergence between the induced human label distribution $\mc{P}$ and the current LLM response preference distribution $\mc{Q}_\phi$: $\mc{L}^{\mathrm{ILHF}}(\phi) = \mathbb{E}_{X_i}\left[\kl{\mc{P}_i}{\mc{Q}_i^\phi}\right]$\footnote{We use the standard convention that $0 \cdot \log(0) = 0$.}. In Section \ref{sec:exps}, we provide an empirical confirmation that fine-tuning via ILHF does indeed yield an inclusive model by converging to the desired response distribution.

\subsection{Efficient Exploration}
\label{sec:efficient-exploration}

While our proposed ILHF loss function can be optimized via traditional supervised-learning techniques, a LLM model can only utilize human feedback for the responses it generates, akin to a reinforcement-learning agent that may only perceive reward signals for the actions executed under its own policy. Given the vastness of the space of possible responses for each prompt, this implies that a LLM model must also contend with the challenge of exploration in its MDP. Fortunately, the reinforcement-learning literature has long-studied the problem of exploration and developed a wide range of solution concepts with varying degrees of statistical efficiency and computational tractability~\citep{kearns2002near,brafman2002r,kakade2003sample,auer2009near,strehl2009reinforcement,jaksch2010near,osband2013more,osband2017posterior,agrawal2017optimistic,jin2018q,lu2019information}. While future work will likely benefit from a deep and meticulous investigation of which concepts from reinforcement learning might fruitfully transfer over to improve the efficiency of LLM fine-tuning, we here offer one concrete suggestion through the use of uncertainty-based exploration. 

Briefly, one principled exploration strategy represents and maintains an agent's epistemic uncertainty~\citep{der2009aleatory} in the underlying MDP or value function and uses it as a quantitative signal to foster exploratory behaviors in a manner that is both provably-efficient~\citep{osband2016deepthesis,osband2016generalization,osband2017posterior,agrawal2017optimistic,osband2019deep,lu2021reinforcement} and computationally-scalable~\citep{osband2016deep,osband2018randomized,o2018uncertainty,dwaracherla2020hypermodels,osband2021epistemic,osband2023approximate}. While the numerous flavors of uncertainty-based explorations schemes also appear with varying degrees of sophistication~\citep{russo2014learning,russo2018learning,lu2021reinforcement}, we leave an investigation of more complex candidates to future work and, instead, focus our attention on those grounded in Thompson sampling~\citep{thompson1933likelihood,russo2018tutorial}, which is both computationally simple and widely used in practice~\citep{li2010contextual,chapelle2011empirical,gopalan2014thompson}. Posterior-sampling methods that employ Thompson sampling for reinforcement learning~\citep{strens2000bayesian,osband2013more,osband2014model,abbasi2014bayesian,agrawal2017optimistic,osband2017posterior,lu2019information} operate in each time period by drawing a single statistically-plausible hypothesis for optimal behavior from the agent's beliefs and proceed to act optimally with respect to the single sample as if it reflects reality. The simplest candidate for maintaining an agent's beliefs and refining them as data accumulates is via ensemble sampling~\citep{lu2017ensemble} which maintains a finite number of randomly-initialized models that can be sampled in each time period and optimized via bootstrapped mini-batches of data~\citep{efron1982jackknife}. Specifically, for each prompt, an Ensemble-ILHF agent samples one model from its ensemble to generate the two responses which induce human feedback. 

\section{Experiment Setup}
\label{sec:exps}

We discuss in this section how we simulate agent-human interactions for all agents used in our empirical evaluation. A reader familiar with reinforcement learning should interpret this section as describing a particular choice of MDP. While language models typically involve a large numbers of tokens in the vocabulary, we will restrict our scope to exactly two: $-1$ and $1$. Typical language models also include a \texttt{STOP} token in the vocabulary that allows for variable-length responses; instead, our synthetic language simply assumes that all response lengths are homogeneous. This is intended to offer a microcosm for studying methods that process and generate tokenized language data. While a reader may feel discouraged at the prospect of results obtained at such a scale orders of magnitude smaller than what is currently driving practical and deployed models in this space, our goal throughout this work is to leverage such simplicity in order to convey maximal clarity. Moreover, if ILHF agents can be shown to bear fruit in such a basic setting, the potential benefits of tackling the same challenges we outline at a larger scale could be far more substantial.

\subsection{A Token-Generating Process \& Pretraining}\label{subsec:datagen}

Consider a synthetic, stateful token-generating process that is governed by a vector $\mu \in \Re^d$, a matrix $W \in \Re^{d \times d}$, and a vector $U \in \Re^d$. The process begins in an initial state $S_0 \in \Re^d$ sampled from a fixed distribution and, at any time $t$, given the state $S_t \in \Re^d$ of this process, a next token $X_{t+1} \in \{-1,1\}$ and state $S_{t+1} \in \Re^d$ are generated according to
\begin{align*}
X_{t+1} &= \left\{\begin{array}{ll}
1 \qquad & \text{w.p. } \frac{\exp\left(\mu^\top S_t\right)}{1 + \exp\left(\mu^\top S_t\right)} \\
-1 \qquad & \text{otherwise}
\end{array}\right., \qquad S_{t+1} = \tanh(W S_t + U X_{t+1}).
\end{align*}
Note that $\tanh$ is applied component-wise.  This generating process can be interpreted as a recurrent neural network with a single hidden layer and a softmax output; indeed, as discussed in the Appendix, all the agents we evaluate adhere to this exact network architecture, only with a larger hidden dimension.  To keep things simple, all prompts and responses will be of a fixed length $\tau \in \mathbb{N}$. One might wonder why this particular token generating process is worth further study. While it is true that there are numerous stateful stochastic processes one could use to model token generation, the one presented above is clearly among the simpler choices while still retaining a sufficient degree of nontrivial structure. 

We offer a preliminary experiment to make the preceding statement precise; namely, that the output token $X_t$ at each time $t$ exhibits long-term dependencies on the history of tokens. Consequently, it suffices for the next output logit, i.e., the probability of the next token being 1, to depend on a relatively long history of logits, since the distribution of the next token is completely determined by the next logit. To demonstrate the dependence, we plot the autocorrelation function of logits in Figure \ref{fig:acorr} (please see the Appendix for the exact autocorrelation formula).

\begin{wrapfigure}{R}{0.4\textwidth}
    \centering
    \includegraphics[width=0.4\textwidth]{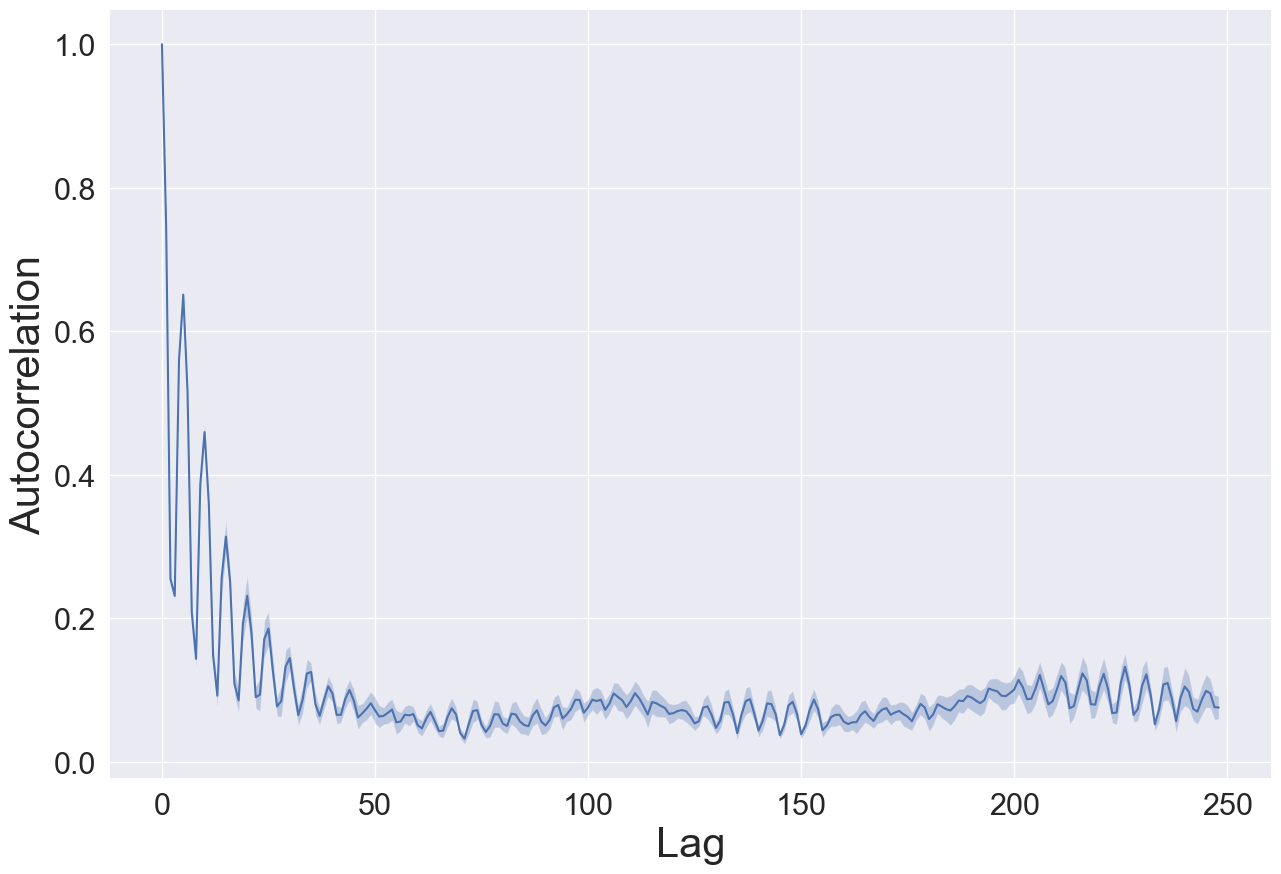}
    \caption{Long-tailed dependence}
    \label{fig:acorr}
    \vspace{-0.5cm}
\end{wrapfigure}


As the goal of our token generating process is to serve as a simplified surrogate to natural language, an important distinction arises between \textit{ideal text} and \textit{shadow text}.  One should think of ideal text as exemplary of well-written text that perfectly aligns with the standards of human evaluators.  This excludes, at the bare minimum, any garbled or malformed token sequences, and more broadly, those that are in discordance with ideal human-level responses; such maligned token sequences are instances of shadow text, (sometimes crude) approximations of ideal text which, for example, appear frequently throughout the Internet and are often intertwined or alongside ideal text. 

We will think of our aforementioned token generating process as one that yields {\it ideal} text so as to be emblematic of proper, linguistically-correct natural language reflective of human preferences.  Modern approaches to pretraining, however, through their reliance on textual data scraped from the Internet, do not rely on text generated by this process but instead on shadow text, an approximation which a downstream agent must use for learning.  For example, the text may not be written by an eloquent writer, may express harmful thoughts, or espouse erroneous responses. Maintaining fidelity to this reality, we assume that this shadow process uses the same matrix $W \in \Re^{d \times d}$ and vector $U \in \Re^d$ but an approximation $\theta \in \Re^d$ of the vector $\mu$.





We conclude this section with a brief discussion of how pretraining with shadow text transpires, clarifying what information all agents in our evaluation will be initialized with at the start of fine-tuning.The pretraining dataset $\mathcal{D}=\{X_{i,1:2\tau}\}_{i=1}^N$ consists of documents generated according to the shadow process.  Each document can be seen as a concatenated prompt-response pair.  Using this dataset, an initial policy $\phi$ is pretrained to maximize the log-probability of predicting the next tokens given the corresponding preceding strings $\min_\phi\mathcal{L}^{\text{pre}}(\phi) = 
 \frac{1}{N} \sum_{i=1}^N \left(\log\pi_\phi(X_{i,1})+\sum_{t=2}^{2\tau}\log\pi_\phi(X_{i,t}|X_{i,1:t-1})\right),$
in a manner that resembles the behavioral cloning~\citep{bain1999framework,pomerleau1989alvinn} algorithm used for imitation learning.  We denote $\phi^{\text{pre}}$ as the policy parameters obtained by pretraining. 

\subsection{Simulating Agent-Human Interactions}

We take each $i$th prompt $X_{i,1:\tau}$ to be sampled independently from our token generating process but with $\theta$ taking the place of $\mu$.  
During training, prompts to language models are often truncated from streams of shadow text; when deployed in practice, one also does not assume that prompts are ideal text free of typographical errors.  These are echoed by our use of the approximate parameter $\theta$ in prompt generation.  Yet another reason that motivates using the shadow process to generate prompts is that, the agent ought to obtain all relevant information about the error $\theta-\mu$ from human feedback.  Thus, prompts should not reveal information about the parameter $\mu$ of the ideal process. 

We consider binary human feedback, where each bit indicates a preference between two responses.  Given the $i$th prompt $X_{i,1:\tau}$, we generate an associated state sequence $S_{i,1:\tau}$ according to $S_{i,t+1} = \tanh(W S_{i,t} + U X_{i,t})$.  Similarly, for each $i$ and $b \in \{0,1\}$, we let $S^{(b)}_{i,0} = S_{i,\tau}$ and generate a state sequence $S^{(b)}_{i,1:\tau}$ associated with response $b$ according to $S^{(b)}_{i,t+1} = \tanh(W S^{(b)}_{i,t} + U X^{(b)}_{i,t})$.  

For each $i$, denote the likelihood function that the ideal text generating process assigns to each response $b \in \{0,1\}$ by
$\ell_{i,b}(\mu) = \prod_{t=1}^\tau\left((1-X^{(b)}_{i,t})\frac{1}{1+\exp(\mu^\top S^{(b)}_{i,t})} + X^{(b)}_{i,t}\frac{\exp(\mu^\top S^{(b)}_{i,t})}{1+\exp(\mu^\top S^{(b)}_{i,t})}\right).$

Then, the binary preference $B_i$ of a random individual is sampled according to $B_i = 
    \begin{cases}
    0 & \text{with probability } \frac{\ell_{i,0}(\mu)}{\ell_{i,0}(\mu) + \ell_{i,1}(\mu)} \\
    1 & \text{with probability } \frac{\ell_{i,1}(\mu)}{\ell_{i,0}(\mu) + \ell_{i,1}(\mu)} 
    \end{cases},$ 
which is the classic Bradley-Terry model for the human preference between response $X^{(1)}_{i,t}$ and $X^{(0)}_{i,t}$. The human feedback we consider is based on the ideal process that employs parameter $\mu$, in spirit with the goal of aligning language model outputs to human preference.  We assume that the human annotator communicates relatively high-quality signals, with the only error stemming from random sampling.  

\subsection{Evaluation Metrics}
\label{subsec:metrics}

A key advantage of studying a simple data-generating process is the ability to, for any prompt generated as described above, tractably compute the KL-divergence between the ideal distribution of responses and distribution of responses produced by any agent as an evaluation metric. Concretely, the ideal process generates a set of independently sampled sequences $\{X_{i,1:2\tau}\}_{i=1}^N$, which can also be viewed as concatenated prompt-response pairs, as well as the corresponding set of state sequences $S_{i,1:2\tau}$. Like the token generating process, a stateful agent then generates corresponding agent state sequences $\{\hat{S}_{i,1:2\tau}\}_{i=1}^N$. For each $i$, let
$\hat{\ell}_i(\mu) = \prod_{t=\tau+1}^{2\tau}\left((1-X_{i,t})\frac{1}{1+\exp(\mu^\top S_{i,t})} + X_{i,t}\frac{\exp(\mu^\top S_{i,t})}{1+\exp(\mu^\top S_{i,t})}\right)$
denote the likelihood function that the ideal text generating process assigns to the $i$-th response, and let $\hat{\ell}_i(\phi)$
denote the likelihood function that the agent's model assigns to the $i$-th response which is identical to $\hat{\ell}_i(\mu)$ with $\phi$ swapped for $\mu$ and $\hat{S}_{i,t}$ in place of $S_{i,t}$. The Monte-Carlo estimation of the KL-divergence can then be expressed as
$\Sc^{\text{KL}}(\phi) = \frac{1}{N}\sum_{i=1}^N \left( \ln \hat{\ell}_{i}(\mu) - \ln \hat{\ell}_{i}(\phi) \right).$
Note that with the token-generating process introduced in Section \ref{subsec:datagen}, there exist an agent that attains zero KL-divergence.  

\subsection{Inclusive Agents}

As a concrete instantiation of our ILHF fine-tuning algorithm, we consider an agent that computes a maximum \textit{a posteriori} (MAP) estimate of $\phi$ at each round of human interaction.  The agent initializes its parameter $\phi_0$ with $\phi^{\text{pre}}$; for the $k$th interaction, it first samples responses from the token generating process with parameter $\phi_k$ and then aims to minimize $\mc{L}^{\mathrm{ILHF}}(\phi)$ which, in the context of our particular token-generating process, simplifies as $\mc{L}^{\mathrm{ILHF}}(\phi) = - \frac{1}{N}\sum_{i=1}^N
\left(\ln \ell_{i,b_i}(\phi) - \ln(\ell_{i,0}(\phi) + \ell_{i,1}(\phi))\right).$
This agent operates in a greedy fashion, using the MAP estimate of parameters to generate responses that humans can subsequently rate.  Note that in our simplified token generating process, we take only one optimization step between interactions, whereas one may engage a significantly larger number of updates in larger models. The full procedure is shown as Algorithm \ref{alg:baseline-log} in the Appendix. 

As an alternative ILHF agent design that leverages exploration strategies to accelerate convergence, we consider a second fine-tuning algorithm, Ensemble-ILHF, that fits an ensemble of models that approximates a posterior distribution.  Before fine-tuning starts, an ensemble of parameters are drawn independently from a normal distribution centered around the pretrained parameter $\phi^{\text{pre}}$. The agent's belief about the variance of the posterior distribution is captured by a covariance matrix $\Sigma$ which, over the course of fine-tuning, is further conditioned on the observed human feedback data. The full procedure for this ensemble agent is shown as Algorithm \ref{alg:ensemble-log} in the Appendix. 

\section{Results and Discussion}
\label{sec:disc}

In this section, we present two sets of computational studies that compare REINFORCE, ILHF, and Ensemble-ILHF agents.  The first set of experiments aim to illustrate how our approach produces an inclusive agent for the didactic example introduced in Section \ref{subsec:inclusive}.  The second set of experiments centers around the token generating process introduced in the previous section, again demonstrating how our fine-tuning procedure yields an inclusive model that captures the desired response distribution while also highlighting the benefits of efficient, uncertainty-based exploration schemes.

In both experiments, all agents follow the same pretraining protocol with 1,000 pretraining samples generated using a shadow process with perturbation variance of 0.3 from the ideal process, yielding identical parameters $\phi^{\mathrm{pre}}$ at the start of fine-tuning for all agents.  All agents use the Adam optimizer~\citep{kingma2014adam} with a learning rate of 0.001 for both pretraining and finetuning.  In each finetuning episode, exactly 64 prompts are provided to the agents to respond to and gain feedback from our synthetic human labels.  All error bars and shaded areas correspond to 1 standard error over 20 seeds. 

\subsection{Didactic Experiment}

\begin{wrapfigure}{R}{0.45\textwidth}
\centering
\includegraphics[width=0.45\textwidth]{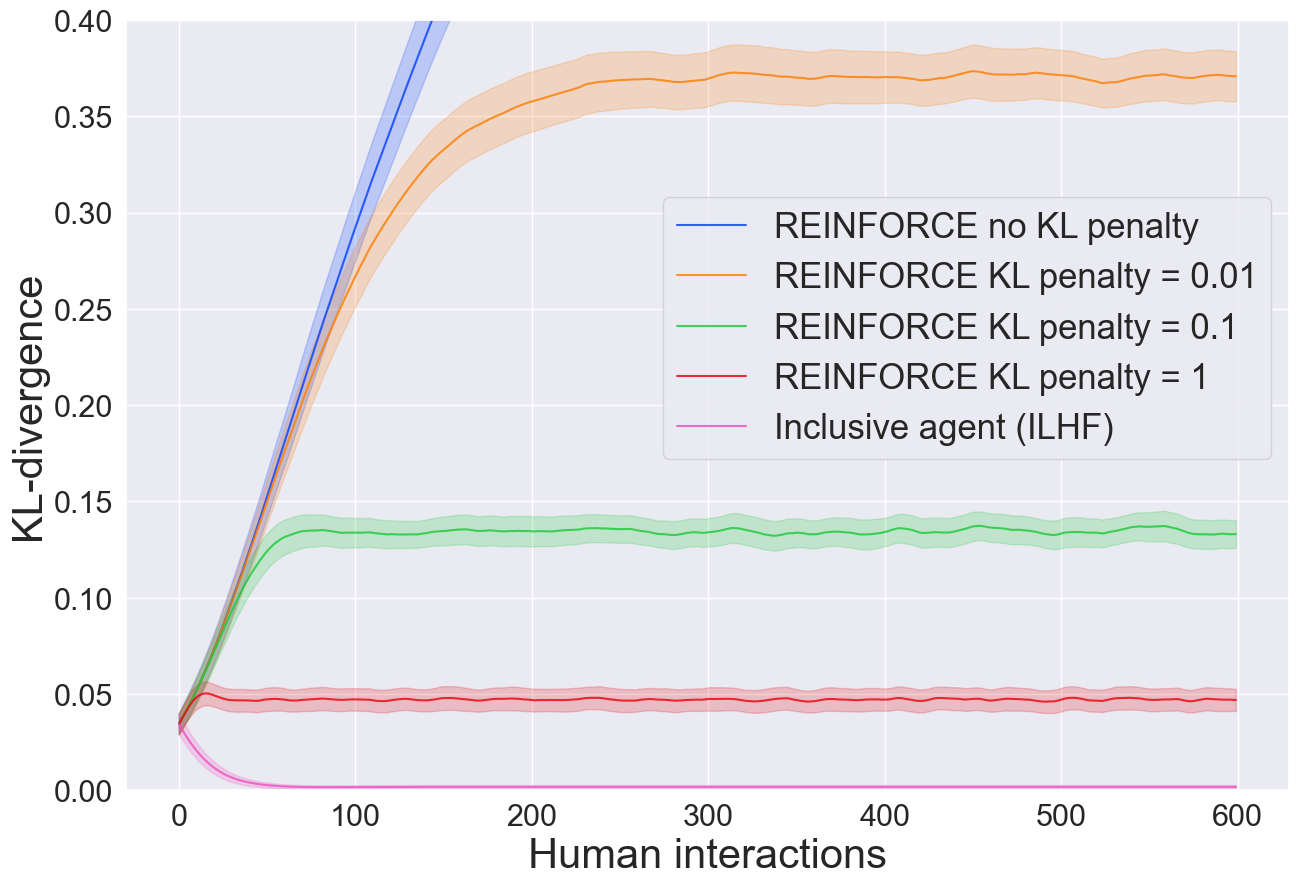}
\captionof{figure}{Reinforce+KL penalty is not inclusive}
\label{fig:toy-inclusive}
\end{wrapfigure}

Figure \ref{fig:toy-inclusive} provides a continuation of the preliminary results shown in Figure \ref{fig:toy-reinforce-rate} for the didactic example of Section \ref{subsec:inclusive} only now showing the KL-divergence metric introduced in Section \ref{subsec:metrics}. 
We first pretrain for 200 epochs before starting the finetuning phase. 
Our KL-divergence metric shows that ILHF is able to learn the ground-truth token distribution, whereas all REINFORCE agents (acting as proxies for the current RLHF paradigm) fail regardless of the KL penalty scale. Notably, the gap between these RLHF agents and our proposed ILHF agent is entirely a function of the divergence between the pretraining and desired response distributions; thus, whenever fine-tuning occurs to correct a significant shift between the two response distributions than what is shown here, the gap between ILHF and RLHF could be significantly larger.

\subsection{Efficient Exploration via Ensemble Sampling}
\label{sec:main-exp}
\begin{figure}[H]
\centering
\begin{tabular}{lr}
\begin{subfigure}{.475\textwidth}
    \centering
    \includegraphics[width=\linewidth]{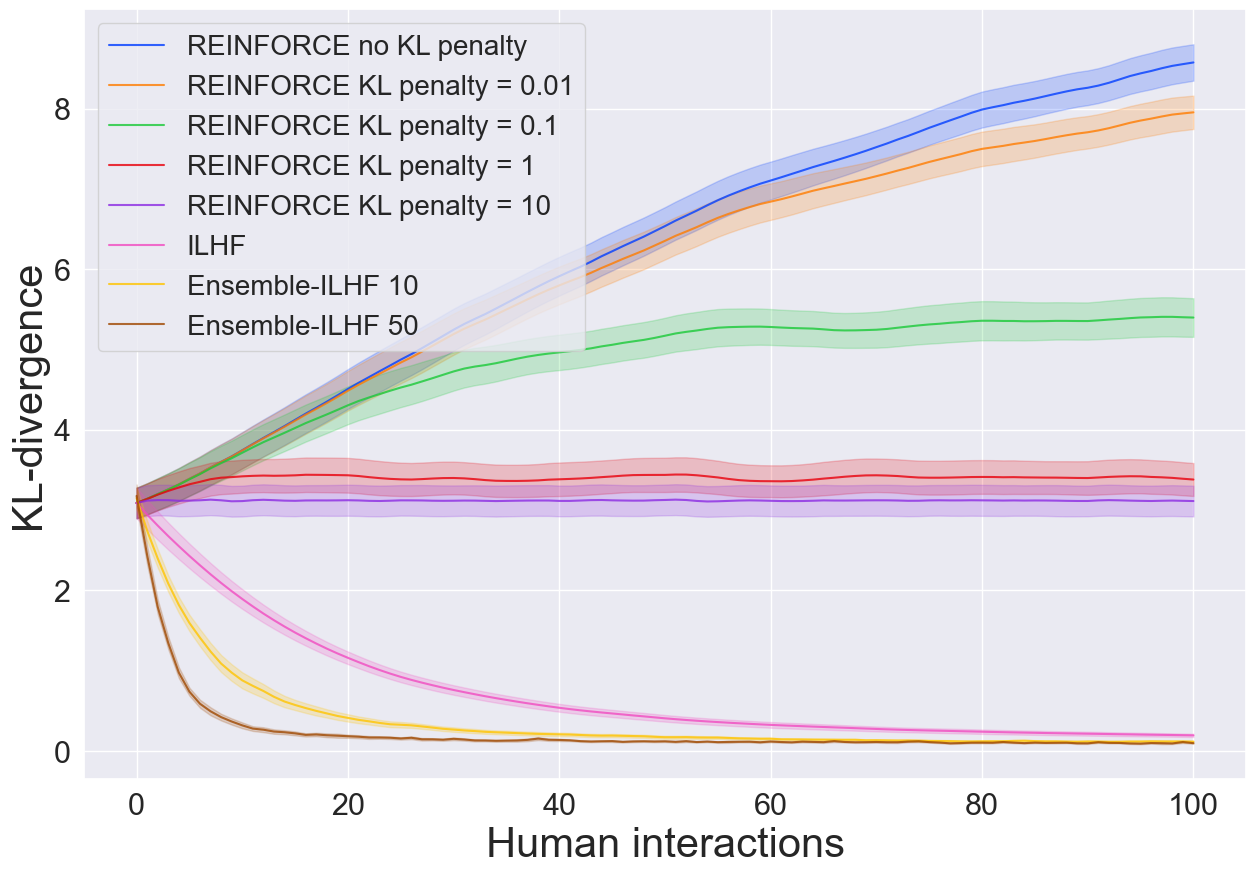}
    \captionof{figure}{KL-divergence vanishes with ILHF but remains large with REINFORCE}
    \label{fig:main-kl}
\end{subfigure}
&
\begin{subfigure}{.475\textwidth}
    \centering
    \includegraphics[width=\linewidth]{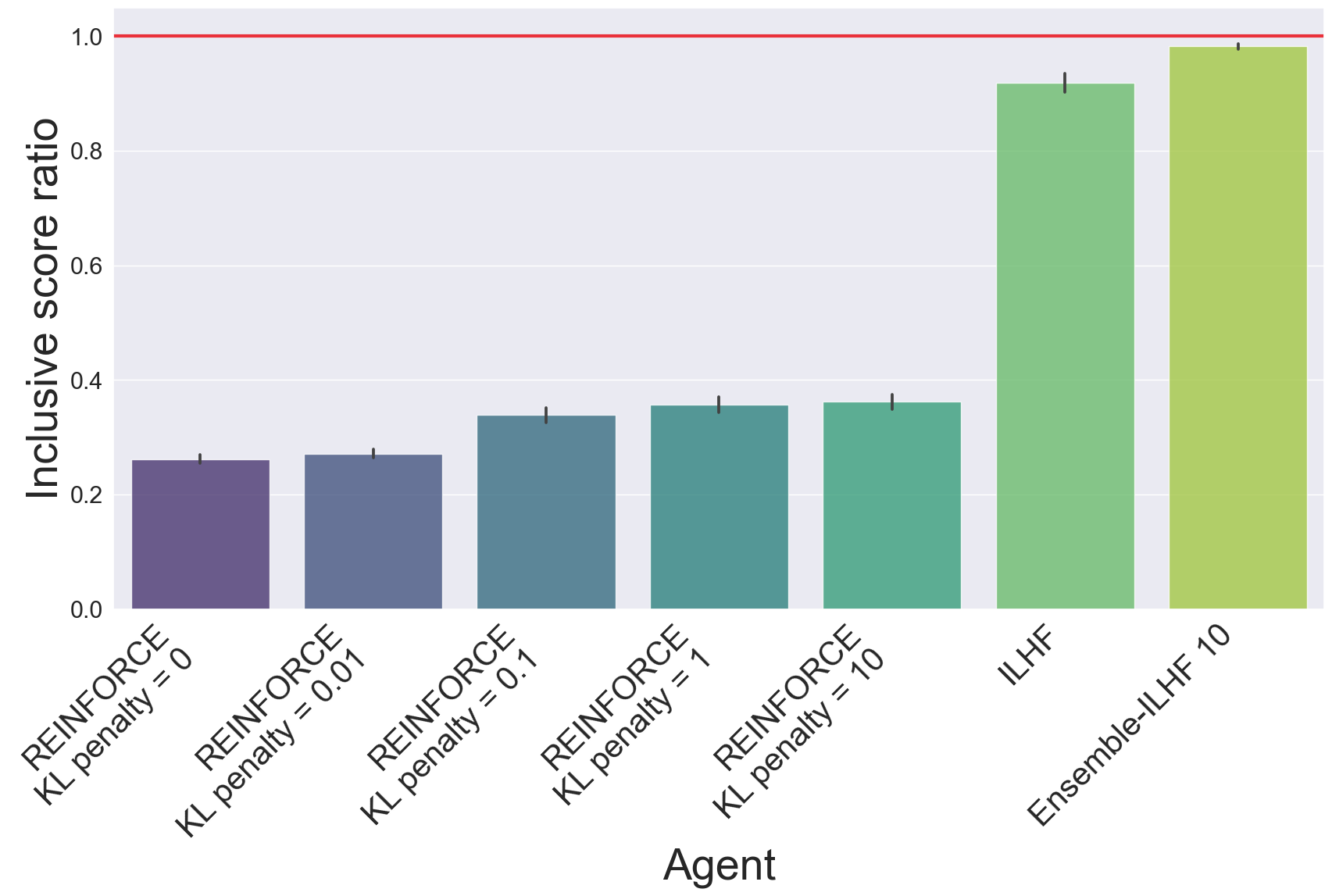}
    \captionof{figure}{Head-to-head comparison of various agents against Ensemble-ILHF 50}
    \label{fig:head-to-head}
\end{subfigure}
\end{tabular}
\label{fig:main_head2head}
\caption{Ensembles improve ILHF, which greatly improves on REINFORCE with KL penalty}
\vspace{-0.3cm}
\end{figure}

Our next set of experiments follows the experiment setup introduced in Section \ref{sec:exps}.  All agents are first pretrained over 20 episodes, then finetuned over 100 episodes of human interactions, as represented by the ideal data generating process.  Note that due to the online nature of the REINFORCE algorithm, during each interaction episode, only one gradient update is performed for each agent model. 
In Figure \ref{fig:main-kl}, we compare the KL-divergence between the response distribution of the ideal process and the response distribution of various agents.
In addition to the KL-penalized REINFORCE agents and ILHF, we also examine the performance of ILHF equipped with an ensemble of models (also called particles) to facilitate exploration, as is introduced in Section \ref{sec:efficient-exploration}.
The figure shows that only ILHF and Ensemble-ILHF can learn the ideal process, while all REINFORCE agents diverge from it regardless of the KL penalty scale.
At the same time, using an ensemble to account for epistemic uncertainty in an inclusive agent significantly accelerates learning.  We provide an ablation study on problem-specific parameters in the Appendix.

Although our synthetic data-generating process allows us to exactly evaluate the KL-divergence between the response distribution of the ideal process and that of an agent, it is not viable to compute in general.  In practice, a head-to-head competition between two agents is typically involved to determine which one performs better.  In a spirit to echo such a procedure, we also carry out a head-to-head competition between our best Ensemble-ILHF agent with 50 particles against all other agents considered in this experiment after finetuning.  Since our goal is to select inclusive agents, instead of employing the agglomerative criterion discussed in Section \ref{subsec:inclusive}, we consider the {\it inclusive score} introduced in \citep{arumugam2022inclusive} which selects the agent that better represents the distribution of human preferences.  For each agent being compared to the Ensemble-ILHF agent with 50 particles, we perform a normalization procedure over the inclusive scores that produces a single statistic, {\it inclusive score ratio}.  The competing agent wins if the ratio is greater than 1 and loses otherwise.  Figure \ref{fig:head-to-head} indicates that Ensemble-ILHF 50 consistently outperforms all other agents in head-to-head combat with statistically significant differences.


\section{Conclusion}

Excitement around the capabilities and prospects of LLMs is a burgeoning area of machine learning, whose prevalence will only continue to grow. Notably, these successes are driven by the RLHF paradigm, which hinges on learning a separate reward model that sits distinct from the LLM itself. To mitigate the challenges of such agglomerative models that collapse towards a single best response, we have proposed Inclusive Learning from Human Feedback (ILHF) as an alternative LLM fine-tuning approach which leverages insights from the field of reinforcement learning to produce inclusive models that preserve the population response distribution. Future work in this area may benefit from incorporating other reinforcement learning ideas to design and optimize LLMs in a more statistically-efficient manner.




\bibliographystyle{plainnat}
\bibliography{references}

\newpage
\appendix

\section{Related Work}
\label{sec:related}

The wealth of knowledge instilled within a generative AI during pretraining is only as good as the procedure developed to judiciously and selectively operationalize that information for downstream tasks of interest. Reinforcement learning~\citep{sutton1998introduction} has emerged as the critical conduit between these pretrained and finetuned models, with dedicated procedures for bridging between them appearing most recently under the name Reinforcement Learning from Human Feedback (RLHF)~\citep{ouyang2022training,huggingface2022rlhf}.

Of course, there is a longstanding history around the incorporation of individual humans into the reinforcement-learning process, spanning a multitude of possible entry points into the underlying formalism~\citep{abel2017agent}, traditionally represented as a Markov Decision Process (MDP)\citep{bellman1957markovian,Puterman94}. While many rich forms of human interaction and intervention are possible, like action labeling~\citep{ross2011reduction,griffith2013policy,ross2014reinforcement} or shared autonomy~\citep{dragan2013policy}, the RLHF paradigm falls in with a broader group of work that tethers human feedback to the reward signal observed by an agent. Preliminary work on this topic exclusively considers human feedback as a exchangeable proxy to the reward signal~\citep{Isbell2001ASR,thomaz2006reinforcement,knox2008tamer,pilarski2011online,knox2012learning,warnell2018deep}, while subsequent work has gone on to question this base premise and explored alternative characterizations through the policy-dependent advantage function~\citep{macglashan2017interactive,arumugam2019deep} or as an indicator of human preferences~\citep{akrour2011preference,wilson2012bayesian,akrour2012april,furnkranz2012preference,akrour2014programming,wirth2016model,el2016score}. With recent progress around generative AIs being driven largely by the capacity for successful deployment of these agents in applications, the latter paradigm has emerged as a default choice for the alignment of complex, monolithic models to human preferences~\citep{christiano2017deep,ibarz2018reward,leike2018scalable,christiano2018supervising,ziegler2019fine,ouyang2022training,bai2022training,glaese2022improving,bakker2022fine}.

\section{Didactic Example}


In an effort to echo the prevailing practice in RLHF, we compare to an alternative baseline, namely the REINFORCE agent \citep{williams1992simple}.  For the $i$-th prompt and its response labeled $b\in\{0,1\}$, the usual Monte-Carlo policy gradient is given by
\begin{equation}\label{eqn:policy-gradient}
\nabla_\phi \ln\ell_{i,b}(\phi)\cdot R_{i,b}, 
\end{equation}
where $R_{i,b}$ is the reward associated with this prompt-response pair.  To keep things simple, we sidestep training a separate reward model, and simply supply the agent with an oracle reward model.  This reward model is represented by the ideal process that produces human feedback. 
 Specifically, the $i$-th prompt and its response labeled $b\in\{0,1\}$ earns reward 
 $$R_{i,b} = \ln\ell_{i,b}(\mu),$$
 where $\ell_{i,b}(\mu)$ is the likelihood function the ideal process assigns to response $b$ given prompt $i$.  The training procedure for this agent is similar to Algorithm \ref{alg:baseline-log}, where the loss function is replaced by 
\begin{align*}
    \Lc^{\text{REINFORCE}}(\phi) &= -\frac{1}{N}\sum_{i=1}^N \left(\ln\ell_{i,0}(\phi)\cdot R_{i,0} + \ln\ell_{i,1}(\phi)\cdot R_{i,1}\right)\\
    &= -\frac{1}{N}\sum_{i=1}^N \left(\ln\ell_{i,0}(\phi)\cdot \ln\ell_{i,0}(\mu) + \ln\ell_{i,1}(\phi)\cdot \ln\ell_{i,1}(\mu)\right).
\end{align*}
Note that the appearance of $\mu$ in the agent's loss function is by design.  Normally, without access to the oracle reward, $R_{i,b}$ would be replaced by the output $R_{i,b}(\psi)$ of a learned reward model separately parameterized by $\psi$. 

A toy example helps demonstrate how this loss function leads to an agglomerative agent, as opposed to the inclusive agent produced by ILHF in Algorithm \ref{alg:baseline-log}.  The example is inspired by the didactic example discussed in Section 2.2 of \cite{arumugam2022inclusive}.  Consider a finetuning dataset consisting of the exact same prompt $X_1=X$, $X_2=X$, \ldots, $X_N=X$.  Suppose each response contains a single token in $\{-1, 1\}$, with $-1$ preferred by two-thirds of the human labelers and $1$ by one-third.  The corresponding logits generating such human feedback are $\{-\ln\sqrt{2}, \ln\sqrt{2}\}$.  While our sequential reward agent only gets to see the 0-1 labels, the REINFORCE agent is endowed with the logits that generate these labels.  As shown in Table \ref{fig:toy-reinforce-rate}, ILHF is able to learn the correct distribution of human labels, whereas REINFORCE over-concentrates onto the preferred response, which is $-1$ in this case.  Notably, this issue is not mitigated by adding a KL-penalty to the policy gradient loss.  

The agent models we use in this example is exactly the same as that explained in Appendix \ref{sec:agent-arch}. 
 In Table \ref{tab:toy-params}, we specify the parameters for this didactic example in our experiments.  
\begin{table}
\captionof{table}{Parameters for the didactic experiment}\label{tab:toy-params}
\centering
\begin{tabular}[b]{ll}
      \toprule
      \textbf{Problem parameter} & \textbf{Value} \\
      \midrule
      prompt $X$ & $[1,\ldots,1]$\\
      \hline
      prompt length & 10\\
      \hline
      evaluation batch size & 3,000\\
      \midrule
      \textbf{Hyperparameter} & \textbf{Value} \\
      \midrule
      ensemble prior scale & 0.01\\
      \hline
      ensemble $L_2$-regularization scale $\eta$ & 1.0\\
      \hline
      agent hidden size $D$ & 2\\
      \bottomrule
\end{tabular}
\end{table}

\section{Token-Generating Process}

Recall that our token-generating process is governed by a vector $\mu\in\Re^d$, a matrix $W\in\Re^{d\times d}$, and a vector $U\in\Re^d$.
Given these parameters and a fixed initial state distribution, sequences are generated in an autoregressive fashion, as defined in Section \ref{subsec:datagen}.  
We instantiate this abstract formulation on specific parameter distributions, and showcase some structural properties implied by this generation process.  

The reason for using a recurrent neural network is so that there is a notion of latent representation that could be useful for predicting the future trajectory.  We choose the underlying data-generating process to be relatively simple, with a low-dimensional latent space.  Specifically, we take $d=2$ in all our subsequent simulations.  The parameters are randomly drawn according to the following distributions:
\begin{itemize}
    \item each entry of the weight matrix $W \in\Re^{2\times 2}$ is drawn i.i.d. from $\mathcal{U}(-\sqrt{\frac{1}{2}},\sqrt{\frac{1}{2}})$,
    \item each entry of the weight vector $U \in\Re^{1\times 2}$ is drawn i.i.d. from $\mathcal{U}(-1,1)$,
    \item the vector $\mu\in\Re^{2}$ is drawn from 2-dimensional standard Gaussian,
    \item the perturbed vector $\theta\in\Re^{2}$ is drawn from $\mathcal{N}(\mu, 0.3\cdot I)$,
    \item the initial hidden state $S_0$ is drawn from 2-dimensional standard Gaussian.
\end{itemize}
Despite the simplicity of this construction, the resulting autoregressive process exhibits nontrivial stationary behavior.  For one, it does not eventually produce all $-1$s or all $1$s or settle on a simple repeated pattern.  For another, each next token depends on a relatively long history of tokens.  To verify that the latter is indeed the case, we consider the autocorrelation of a series of logits. 
Let $(a_i)_{i=1}^\tau$ denote the logits at the last layer of the model for $\tau$ outputs.  For each $j,k\in\{1,\ldots,\tau\}$ such that $j<k$, let
$$\overline{a_{j:k}} = \frac{1}{k-j}\sum_{i=j}^k a_i$$
denote the mean of the logits $a_j,a_{j+1},\ldots,a_k$.  For $k\ge 0$, the autocorrelation function at lag $k$ is given by
\begin{align*}
    \alpha(k) &= \frac{1}{\tau-k} \sum_{i=1}^{\tau-k} \frac{(a_i - \overline{a_{1:\tau-k}}) (a_{i+k} - \overline{a_{k+1:\tau}})}{\sqrt{\sum_{i=1}^{\tau-k} (a_i - \overline{a_{1:\tau-k}})^2 \sum_{i=1}^{\tau-k} (a_{i+k} - \overline{a_{k+1:\tau}})^2 }}.
\end{align*}
We plot the function $\alpha$ against lag in Figure \ref{fig:acorr}.

\section{Agent Designs}\label{sec:agent-arch}

\subsection{Agents}
In Section \ref{subsec:datagen}, we introduced the architecture for the token-generating process in our experiments. Our agents are equipped with similar architectures as the token generating process, but with high dimensional hidden states.  To distinguish learned from data generating models, recall that we denote the parameters of the latter by $\theta\in\Re^d$, $W\in\Re^{d\times d}$, and $U\in\Re^d$, whereas we denote the agent's parameters by a vector $\phi\in\Re^D$, a matrix $\hat{W}\in\Re^{D\times D}$, and a vector $\hat{U}\in\Re^D$.  As before, the process begins in an initial state $\hat{S}'_0\in\Re^D$ sampled from a fixed distribution.  Given a prompt $X_{1:\tau}$, the agent generates an associated agent state sequence $\hat{S}'_{1:\tau}$ according to $\hat{S}'_{t+1} = \tanh(\hat{W} \hat{S}'_t+\hat{U} X_t)$.  We let $\hat{S}_0 = \hat{S}'_\tau$. Then at any time $t$, given the state $\hat{S}_t\in\Re^D$ of this process, a next token $\hat{X}_{t+1}\in\{-1,1\}$ and state $\hat{S}_{t+1}\in\Re^D$ are generated according to
\begin{align*}
\hat{X}_{t+1} &= \left\{\begin{array}{ll}
1 \qquad & \text{w.p. } \frac{e^{\phi^\top \hat{S}_t}}{1 + e^{\phi^\top \hat{S}_t}} \\
-1 \qquad & \text{otherwise.}
\end{array}\right. \\
\hat{S}_{t+1} &= \tanh(\hat{W} \hat{S}_t + \hat{U} \hat{X}_{t+1}).
\end{align*}
A more complex agent model than the true data generating process affords the agent to a rich latent state representation that is likely to encode a close approximation to the true latent state, which can be useful when fine-tuning for various downstream tasks.  The agent's larger model grants it the ability to fit to the data generating process almost perfectly without advanced knowledge of the latent features driving it.  This is akin to the manner in which overparameterized neural networks are able to discover latent spaces.  

All our agents follow the same pretraining procedure described in Section \ref{subsec:datagen} that fits a model to the pretraining data to produce a point estimate, which is then fine-tuned using the human feedback data.  Thus, before fine-tuning, our agents are endowed with knowledge learned in pretraining, represented by $\hat{W}$, $\hat{U}$, and $\phi^{\text{pre}}$.  During fine-tuning, an agent progresses over rounds of human interaction.  Each round begins with the agent observing a prompt.  The agent then produces two responses, which could depend on the knowledge from pretraining as well as human feedback received over previous rounds.  Finally, the agent observes an indication of preference provided by a random individual, and moves on to the next round.  The problem parameters and agent hyperparameters used throughout the experiments are presented in Table \ref{tab:agent-params}. 

\begin{table}
\centering
\begin{tabular}[b]{ll}
      \toprule
      \textbf{Problem parameter} & \textbf{Value} \\
      \midrule
      prompt length $\tau$ & 64\\
      \hline
      softmax temperature & 3.0\\
      \hline
      evaluation batch size & 500\\
      \midrule
      \textbf{Hyperparameter} & \textbf{Value} \\
      \midrule
      ensemble prior scale & 0.01\\
      \hline
      ensemble $L_2$-regularization scale $\eta$ & 1.0\\
      \hline
      agent hidden size $D$ & 10\\
      \bottomrule
\end{tabular}
\captionof{table}{Parameters for agents}\label{tab:agent-params}
\end{table}

\clearpage
\section{Algorithms}
\label{sec:algs}

\begin{algorithm}[H]
\caption{Inclusive Learning from Human Feedback (ILHF)}
\begin{tabular}{p{1cm}ll}
{\bf Inputs:} 
&parameters     &pretrained parameters $\hat{W}, \hat{U}, \phi^{\text{pre}}$\\
&data           &prompt generator $\mathtt{PromptGen}(\tau)$, batch size $N$\\
&human labeler  &$\mathtt{Human(\cdot,\cdot)}$, number of interaction episodes $K$\\
&loss function  &$\Lc^{\text{ILHF}}(\phi; \Dc)$ evaluates parameter $\phi$ on dataset $\Dc$\\
&optimization   &update rule $\mathtt{Optimizer}$, minibatch loader $\mathtt{DataLoader}$\\
{\bf Returns:}  &$\phi_K$ &parameters for the trained language model
\end{tabular}
\begin{algorithmic}[1]
\State {\bf Initialize:} initial agent parameters $\phi_0 \gets \phi^{\text{pre}}$
\For {epoch $k=0,\dots,K$}
\For {$i=1,\ldots,N$}
\State sample prompt $x_{i,1:\tau} \gets\mathtt{PromptGen}(\tau)$
\State generate response pair $(x_{i,1:\tau}^{(0)},x_{i,1:\tau}^{(1)})$
\State observe human label $b_i\gets \mathtt{Human}(x_{i,1:\tau}^{(0)},x_{i,1:\tau}^{(1)})$
\EndFor
\State $\Dc^k\gets (x_{i,1:\tau}, x_{i,1:\tau}^{(0)},x_{i,1:\tau}^{(1)}, b_i)_{i=1}^N$
\For {minibatch $\Dc$ in $\mathtt{DataLoader}(\Dc^k)$}
\State compute $\mathtt{gradient}\gets \nabla_{\phi|\phi=\phi_k} \Lc^{\text{ILHF}}(\phi; \Dc)$
\State update $\phi_{k+1}\gets \mathtt{Optimizer}(\phi_k, \mathtt{gradient})$
\EndFor
\EndFor
\end{algorithmic}\label{alg:baseline-log}
\end{algorithm}

\begin{algorithm}[H]
\caption{ILHF with Ensemble Models (Ensemble-ILHF)}
\begin{tabular}{p{1cm}ll}
{\bf Inputs:}
&parameters     &pretrained parameters $\hat{W}, \hat{U}, \phi^{\text{pre}}$\\
&data           &prompt generator $\mathtt{PromptGen}(\tau)$, batch size $N$\\
&ensemble       &ensemble size $M$, prior scale $\Sigma$, prior regularization scale $\eta$\\
&human labeler  &$\mathtt{Human(\cdot,\cdot)}$, number of interaction episodes $K$\\
&loss function  &$\Lc^{\text{ILHF}}(\phi; \Dc)$ evaluates parameter $\phi$ on dataset $\Dc$\\
&optimization   &update rule $\mathtt{Optimizer}$, minibatch loader $\mathtt{DataLoader}$\\
{\bf Returns:}  &$(\phi_{K,m})_{m=1}^M$ & ensemble parameters for the trained language model
\end{tabular}
\begin{algorithmic}[1]
\State {\bf Initialize:} initial ensemble parameters $\phi_{0,m} \gets \mathcal{N}(\phi^{\text{pre}}, \Sigma)$, $m=1,\ldots,M$
\For {epoch $k=0,\dots,K$}
\For {$i=1,\ldots,N$}
\State sample prompt $x_{i,1:\tau} \gets\mathtt{PromptGen}(\tau)$
\State sample ensemble index $z_i\sim\text{Unif}\{1,\ldots,M\}$
\State generate response pair $(x_{i,1:\tau}^{(0)},x_{i,1:\tau}^{(1)})$ using $\phi_{k,z_i}$
\State observe human label $b_i\gets \mathtt{Human}(x_{i,1:\tau}^{(0)},x_{i,1:\tau}^{(1)})$
\EndFor
\State $\Dc^k\gets (x_{i,1:\tau}, x_{i,1:\tau}^{(0)},x_{i,1:\tau}^{(1)}, b_i)_{i=1}^N$ 
\For {$m=1,\ldots,M$}
\For {minibatch $\Dc$ in $\mathtt{DataLoader}(\Dc^k)$}
\State randomly perturb $\Dc$ (double or nothing)
\State compute $\mathtt{gradient}\gets \nabla_{\phi|\phi=\phi_{k,m}} \Lc^{\text{ILHF}}(\phi;\Dc) + \eta\|\phi - \phi_0\|_2^2$
\State update $\phi_{k+1,m}\gets \mathtt{Optimizer}(\phi_{k,m}, \mathtt{gradient})$
\EndFor
\EndFor
\EndFor
\end{algorithmic}\label{alg:ensemble-log}
\end{algorithm}

\clearpage
\section{Ablations}
In addition to the experiments in the main paper, we perform ablation studies on problem parameters and agent hyperparameters.  All error bars and confidence intervals are 1 standard error over 20 seeds. 

\begin{figure*}[htbp]
  \centering
  \begin{minipage}[t]{0.45\textwidth}
    \centering
    \includegraphics[width=\linewidth]{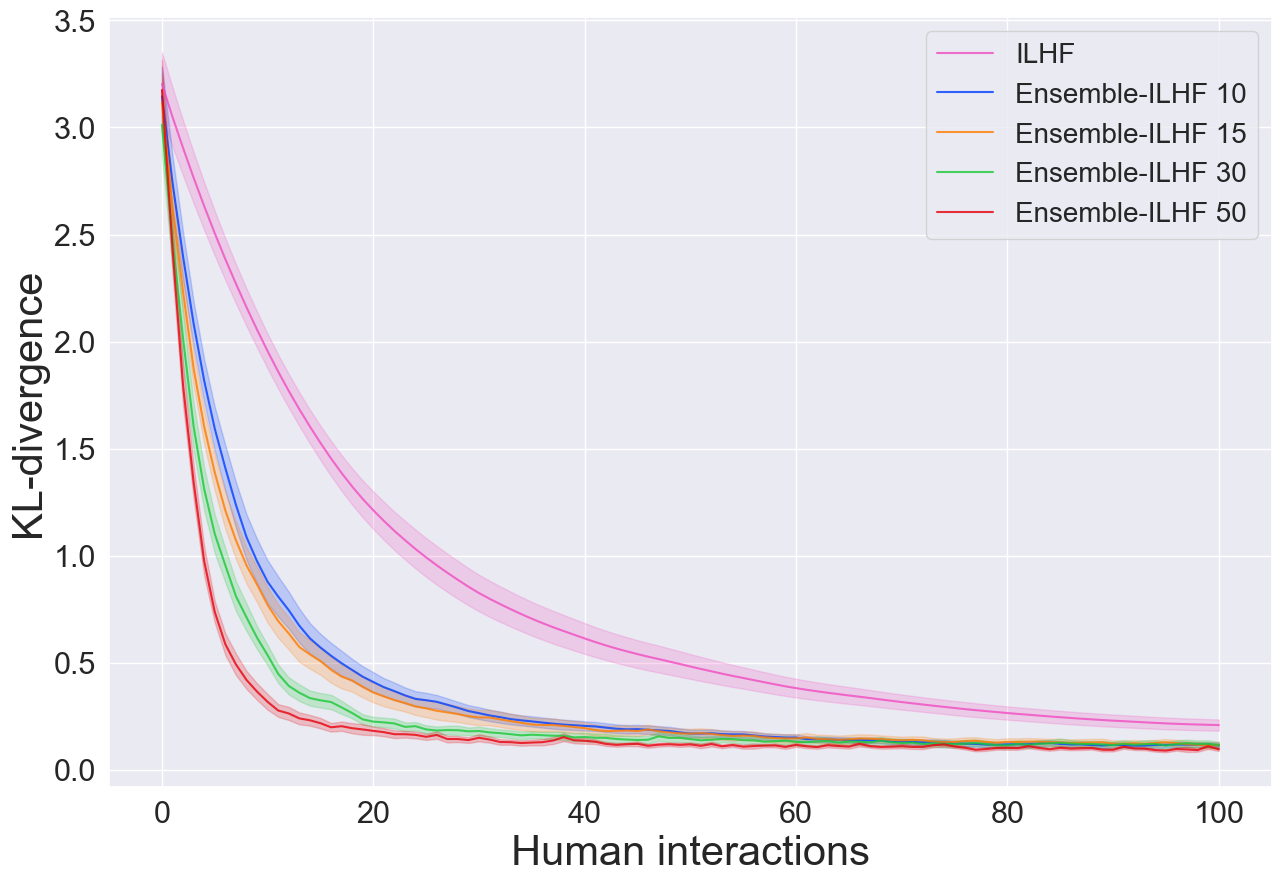}
    \captionof{figure}{ILHF with ensembles converges faster as the number of particles increases.}
    \label{fig:ablate-particles}
  \end{minipage}
  \hfill
  \begin{minipage}[t]{0.45\textwidth}
    \centering
    \begin{tabular}[b]{ll}
      \toprule
      \textbf{Agent} & \textbf{Inclusive score ratio} \\
      \midrule
      Ensemble-ILHF 10 & $1.0384\pm 0.0061$\\
      \hline
      Ensemble-ILHF 15 & $1.0467\pm 0.0084$\\
      \hline
      Ensemble-ILHF 30 & $1.0338\pm 0.0062$\\
      \hline
      Ensemble-ILHF 50 & $1.0365\pm 0.0088$\\
      \bottomrule
    \end{tabular}
    \captionof{table}{All Ensemble-ILHF agents beat ILHF.}\label{tab:incl-64}
  \end{minipage}
\end{figure*}

\paragraph{Ensemble size ablations.} We further vary the number of ensembles in Ensemble-ILHF and plot the KL-divergence in Figure \ref{fig:ablate-particles}.  As we increase the number of ensemble particles, the convergence speed increases, but the marginal gain decreases.  Table \ref{tab:incl-64} shows the inclusive score ratio, a metric explained in Section \ref{sec:main-exp}, of different Ensemble-ILHF agents compared against a ILHF agent after 100 human interactions.

\begin{figure*}[htbp]
  \centering
  \begin{minipage}[t]{0.45\textwidth}
    \centering
    \includegraphics[width=\linewidth]{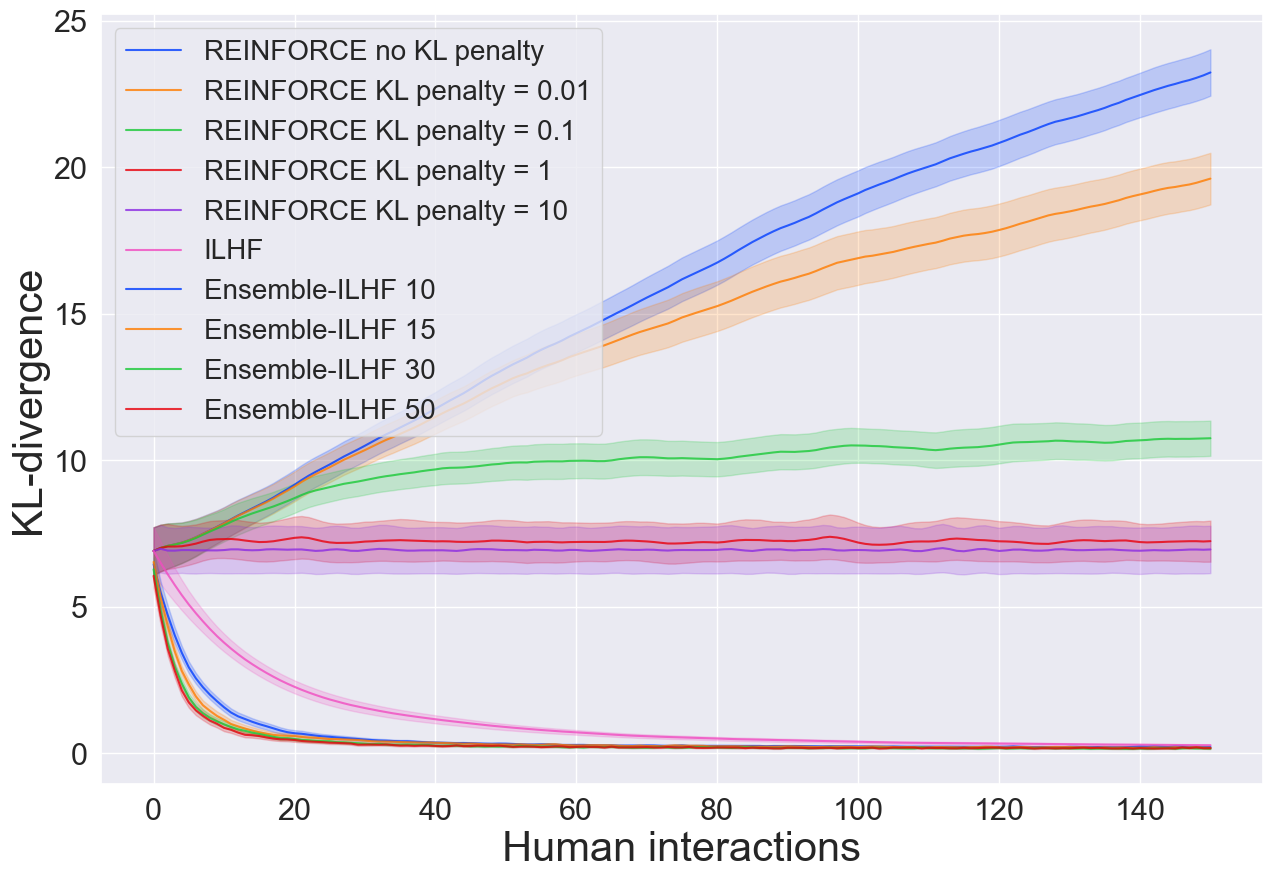}
    \captionof{figure}{Comparisons of REINFORCE with various KL penalty scales, ILHF, and Ensemble-ILHF with various ensemble sizes when $\tau=128$.}
    \label{fig:ablate-128}
  \end{minipage}
  \hfill
  \begin{minipage}[t]{0.45\textwidth}
    \centering
    \includegraphics[width=\linewidth]{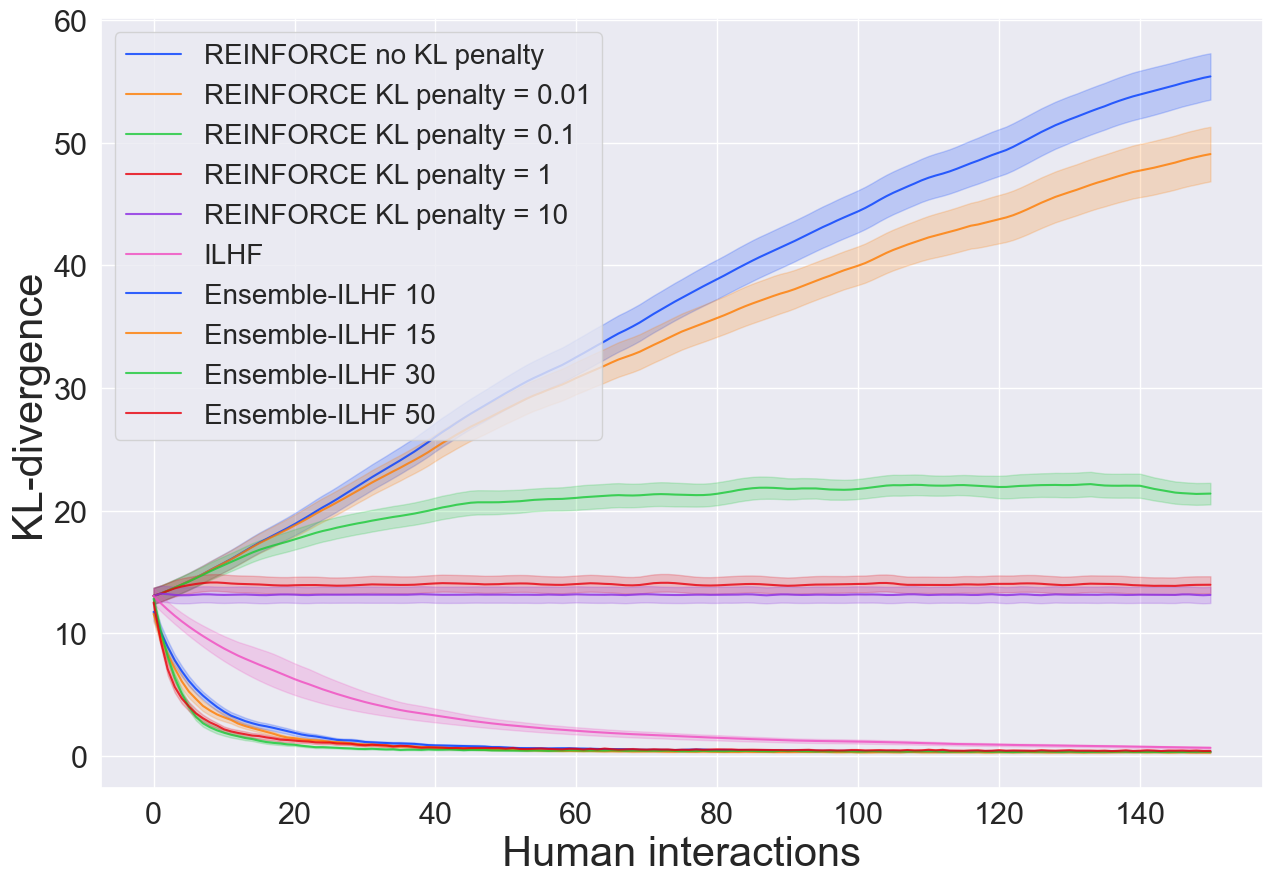}
    \captionof{figure}{Comparisons of REINFORCE with various KL penalty scales, ILHF, and Ensemble-ILHF with various ensemble sizes when $\tau=256$.}
    \label{fig:ablate-256}
  \end{minipage}
\end{figure*}

\paragraph{Sequence length ablations.} We vary the sequence length from $\tau=64$ to $\tau\in\{128, 256\}$ in our experiments and observe similar qualitative behavior, as shown in Figures \ref{fig:ablate-128} and \ref{fig:ablate-256}.  Since increasing $\tau$ in general requires a longer horizon for the agents to learn, we also increase the number of human interactions to $150$ for these ablations.  

\end{document}